\pdfoutput=1

\documentclass[11pt]{article}

\usepackage[final]{acl}

\usepackage{times}
\usepackage{latexsym}

\usepackage{amsmath}
\usepackage[normalem]{ulem}
\usepackage[T1]{fontenc}

\usepackage[utf8]{inputenc}

\usepackage{microtype}

\usepackage{inconsolata}

\usepackage{graphicx}
\usepackage{enumitem}
\usepackage{cleveref}
\usepackage{array}
\usepackage{booktabs}
\usepackage{algorithm}
\usepackage{algpseudocode}
\usepackage{amssymb}
\usepackage{amsmath}
\usepackage{tcolorbox} 
\usepackage{multirow}
\usepackage{xcolor}
\usepackage{amsmath}
\usepackage{xcolor,pifont}
\usepackage[dvipsnames]{xcolor}
\usepackage{tabularx}
\usepackage{makecell}

\usepackage[table]{xcolor}   
\usepackage[dvipsnames]{xcolor}
\usepackage{graphicx}
\usepackage{caption}

\usepackage{booktabs}
\usepackage[table]{xcolor}

\newcommand*\colourcheck[1]{%
  \expandafter\newcommand\csname #1check\endcsname{\textcolor{#1}{\ding{52}}}%
}
\colourcheck{blue}
\colourcheck{green}
\colourcheck{red}

\title{
   Reinforcing Step-level Reasoning for Effective Self-Correction in LLMs
}

\author{
    Vu Duc Anh$^{1}$ \quad Nhat M. Hoang$^{1}$ \quad Do Xuan Long$^{2, 4}$ \\ \textbf{Cong-Duy Nguyen$^{3}$ \quad Ponhvoan Srey$^{1}$ \quad Luu Anh Tuan$^{1,3}$\thanks{~~Corresponding author.}}\\
    $^{1}$Nanyang Technological University, Singapore  \quad
    $^{2}$National University of Singapore \\
    $^{3}$VinUniversity, Vietnam  \quad
    $^{4}$Institute for Infocomm Research (I$^2$R), A*STAR\\
    \texttt{\{vuducanh001, hoangmin003, ponhvoan002, anhtuan.luu\}@ntu.edu.sg}, \\
    \texttt{xuanlong.do@u.nus.edu}  \quad \texttt{duy.ntc@vinuni.edu.vn
}\\
}

\begin{document}
\maketitle
\begin{abstract}
Achieving effective self-correction, where models verify and correct their own mistakes, remains a fundamental challenge for large language models (LLMs). In this work, we propose Self-Fix Step-DPO (SFS-DPO), a reinforcement learning based, two-stage framework for step-level self-verification and self-correction. The first stage strengthens step-level reasoning via step-level preference optimization, while the second stage explicitly trains models to self-verify and self-correct. We further introduce a teacher-assisted variant, SFS-DPO-R, which incorporates explanatory rationales for error verification to provide stronger corrective signals. Comprehensive in-domain and out-of-domain evaluations across multiple LLMs demonstrate that SFS-DPO and SFS-DPO-R consistently outperform prior step-level training baselines. Our analysis further reveals improvements in self-correction frequency and effectiveness, highlighting the importance of strengthening step-level reasoning for robust performance. 
\end{abstract}

\begin{figure}[t!]
\centering
\includegraphics[width=0.9\columnwidth]{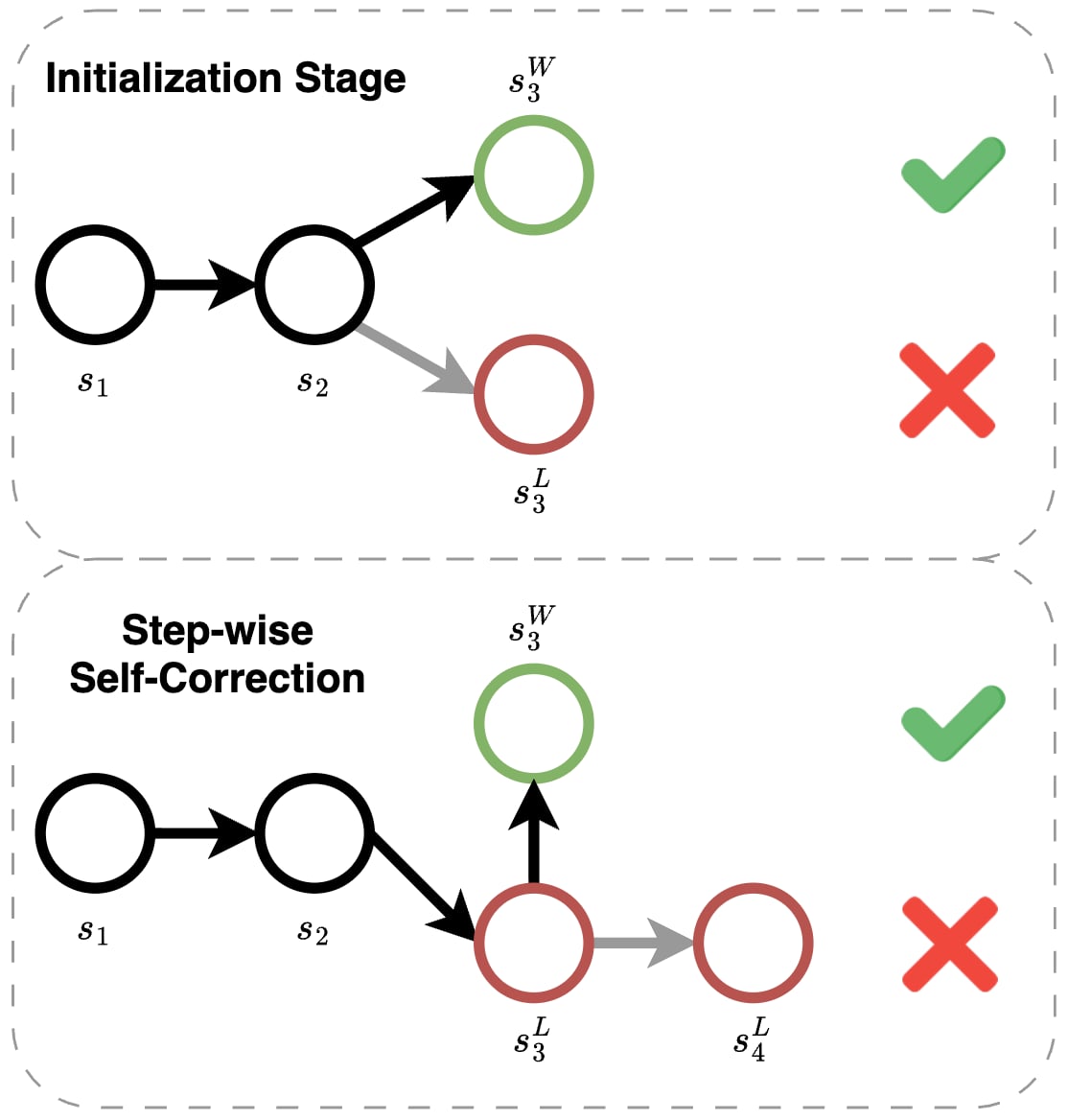}
\caption{
    \textbf{Initialization Stage (Stage 1)} learns local step preferences by favoring correct continuations over incorrect ones, without revising errors. \textbf{Step-wise Self-Correction (Stage 2)} complements this by training explicit self-corrections: revising incorrect steps toward preferred continuations before proceeding, converting step-wise preferences into targeted multi-step reasoning improvements.
} 
\label{fig:method}
\end{figure}

\section{Introduction}
\label{sec:intro}





{Despite recent advances in frontier large language models (LLMs), smaller LLMs still struggle with complex math reasoning}, where early errors can propagate throughout the entire solution \cite{lightman2024lets, hong-etal-2024-orpo, chen2025towards}. To provide more localized learning signals, recent studies have shifted toward step-level objectives \cite{lai2024step, lu2024step, xu2025full, pham2026grace}. However, they mainly optimize preferences over better reasoning continuations, without explicitly correcting erroneous steps that were already generated \cite{kumar2024training, pan-etal-2025-lemma}.


\begin{table*}[t]
\centering
\label{tab:self-correction-baselines}
\resizebox{0.99\textwidth}{!}{
\begin{tabular}{lcc r@{/}l cccc}
\toprule
\textbf{Method} 
& \textbf{Init.} 
& \textbf{Optimization} 
& \multicolumn{2}{c}{\makecell{\textbf{Training Size}\\\textbf{Init.}\,/\,\textbf{Opt.}}}
& \textbf{Spont.} 
& \textbf{Err. Detect.}
& \textbf{Error Critic}
& \textbf{External Teacher} \\
\midrule

SCoRe
& RL
& RL
& 12K&12K 
& \textcolor{red}{\ding{55}}
& \textcolor{red}{\ding{55}}
& \textcolor{red}{\ding{55}}
& \textcolor{red}{\ding{55}}
\\

SuperCorrect
& SFT
& Step RL
& 100K&10K
& \greencheck  
& \greencheck 
& \greencheck         
& \greencheck
\\

LEMMA
& -
& SFT
& -&88.9K
& \greencheck  
& \greencheck 
& \textcolor{red}{\ding{55}}        
& \textcolor{red}{\ding{55}}
\\

S3C-MATH
& -
& SFT
& -&927K
& \greencheck 
& \greencheck 
& \greencheck         
& \greencheck
\\

SPOC
& SFT
& RL
& 860K&-
& \greencheck  
& \greencheck 
& \greencheck         
& \textcolor{red}{\ding{55}}
\\

S$^2$R
& SFT
& RL
& 3.1K&10K 
& \greencheck
& \greencheck 
& \greencheck         
& \textcolor{red}{\ding{55}}
\\

\midrule

SFS-DPO (ours)
& Step RL
& Step RL
& 10K&8.4K
& \greencheck  
& \greencheck 
& \textcolor{red}{\ding{55}}        
& \textcolor{red}{\ding{55}}
\\

SFS-DPO-R (ours)
& Step RL
& Step RL
& 10K&8.4K 
& \greencheck  
& \greencheck 
& \greencheck         
& \greencheck
\\

\bottomrule
\end{tabular}}
\caption{{Comparison of self-correction methods in LLMs. Our methods (SFS-DPO and SFS-DPO-R) achieve competitive capability coverage with significantly less training data than prior approaches.}}
\end{table*}

This limitation motivates a complementary paradigm of \emph{self-correction}, which introduces self-improving loops that enable models to detect and correct their own mistakes during inference. This has inspired a parallel line of research on training LLMs to self-correct
\citep{kumar2024training,yang2025supercorrect,zhao2025boosting}.
While these methods highlight the promises of self-correction, they introduce optimization challenges. 
Specifically, \citet{kumar2024training} demonstrate that supervised fine-tuning (SFT) on correction traces is prone to distribution shift and behavior collapse, and propose an on-policy reinforcement learning (RL) solution. To improve the self-correction robustness, \citet{ma2025s,yang2025supercorrect} further decompose the optimization process into two phases: first, training the models using SFT with carefully designed templates, followed by a second RL phase for self-correction. However, these initialization strategies are primarily designed to enforce models to follow specific templates that facilitate the self-correction phase, rather than to explicitly strengthen step-level reasoning. Explicitly strengthening step-level reasoning is crucial, since step-wise self-correction involves two challenging subproblems, error detection and targeted revision, and weak step-level reasoning can cause their joint learning to produce noisy signals and compounding errors \citep{caruana1997multitask}.

We introduce Self-Fix Step-DPO (SFS-DPO), a two-stage framework that trains LLMs to self-verify and self-correct via {step-level} reinforcement learning. As shown in \Cref{fig:method}, the initialization stage applies an RL-based step-level preference optimization to strengthen step-wise reasoning, while the second stage (Step-wise Self-Correction) trains the models to self-correct via learning the preferences between a self-corrected continuation and the continuation produced when an incorrect step is left unaddressed. We further propose a teacher-assisted variant, SFS-DPO-R, which augments corrections with explanatory rationales to provide stronger corrective signals and improve downstream accuracy. Comprehensive in-domain and out-of-domain evaluations on multiple LLMs show that SFS-DPO and SFS-DPO-R yield consistent gains over baselines. Our contributions are threefold:

\begin{itemize}
    \item We propose an RL-based, step-level self-correction training framework with two variants, SFS-DPO and SFS-DPO-R, that improves step-level reasoning and enables effective self-correction in LLMs.

    \item We conduct comprehensive in-domain and out-domain experiments demonstrating that our method achieves consistent improvements over existing baselines across multiple LLMs.

    \item We analyze self-correction behavior in our framework, providing empirical insights into its frequency, effectiveness, and relationship with reasoning performance.

\end{itemize}

\section{Related Work}

\subsection{RL for LLM Reasoning}

Preference optimization has been widely used to align language models for reasoning tasks, typically operating at the level of full solutions or complete responses \citep{rafailov2023direct}. However, answer-level preferences can be coarse for complex math reasoning, as incorrect solutions may share long correct prefixes with correct ones, leading to weak credit assignment. To address this limitation, recent work has explored learning signals at the level of individual reasoning steps \citep{chen-etal-2024-step,lai2024step,xu2025full,lightman2024lets,jin2025recut,lahlou2025port}. Notably, Step-DPO \citep{lai2024step} defines preference comparisons over next-step continuations conditioned on a correct prefix, explicitly targeting the first erroneous step in a reasoning trajectory and providing localized supervision without requiring explicit step-level labels. In addition, process supervision \citep{lightman2024lets} demonstrates that supervising intermediate reasoning steps with explicit correctness judgments significantly improves mathematical reasoning by enabling more precise credit assignment. Our work builds on this line by using step-level preference optimization as a first training stage for learning step-wise self-correction.

\subsection{Teaching LLMs to Self-Correct}
Self-correction is a critical capability for scaling the reliability of large language models. As a result, a growing body of work studies how to equip LLMs to self-correct their own reasoning \citep{kumar2024training,do2024multi,yang2025supercorrect,zeng2025evolving,zhao2025boosting}, see \Cref{tab:self-correction-baselines}. 
Notably, SCoRe \citep{kumar2024training} learns intrinsic self-correction via multi-turn on-policy reinforcement learning on self-generated data, highlighting distribution shift and behavior collapse as limitations of offline correction training. SuperCorrect \citep{yang2025supercorrect} proposes a two-stage teacher--student framework that initializes models with supervised fine-tuning on hierarchical thought templates and then applies preference optimization using teacher-provided correction traces. S3C-MATH \citep{yan2025s} trains step-level self-correction by inserting incorrect steps into correct solution traces and supervising models to detect and fix these errors during generation. SPOC \citep{zhao2025boosting} induces spontaneous self-correction at inference time by alternating solution generation and verification using reinforcement learning. S$^2$R \citep{ma2025s} combines supervised initialization with outcome- and process-level reinforcement learning to train models that self-verify and iteratively refine their reasoning. 
In contrast, our work initializes step-level self-correction with a step-wise RL-based preference optimization stage, which we empirically find yields more robust downstream self-correction than from scratch or supervised fine-tuning.




\section{Methodology}

\paragraph{Task Formulation.} We formulate self-correction in language model reasoning as the generation of a multi-step trajectory with explicit error detection and repair. Given an input problem $x$, a language model $\pi_\theta$ produces a reasoning trajectory
\[
\{ s_j \}_{j=1}^{M} = (s_1, \ldots, s_M, \hat{y}),
\]
where each step $s_j$ is generated autoregressively as
\[
s_j \sim \pi_\theta(\cdot \mid x, s_{<j})
\]
We formulate self-correction by allowing each step $s_j$ to take one of three types,
$\{\textsc{solution step}$, $\textsc{error-detection step}$, $\textsc{fixed step}\}$,
corresponding respectively to a standard reasoning step, an explicit signal (optionally with reasoning) indicating that the previous step is incorrect, or a corrected version of an erroneous step. When an \textsc{error-detection step} is  generated, the model produces a \textsc{fixed step} that replaces the incorrect reasoning in the trajectory. The final answer $\hat{y}$ is obtained from the terminal corrected trajectory.

\paragraph{Motivation.} Step-wise self-correction requires optimizing two objectives simultaneously: evaluating the correctness of a reasoning step and generating a corrected alternative. Jointly optimizing these objectives can produce noisy learning signals when step-level reasoning capability is insufficient, as errors in step evaluation and step revision may compound \citep{caruana1997multitask}. Our framework, therefore, adopts a two-phase training strategy: (1) \textbf{Step-level Preference Optimization} as a initialization stage that improves step-level reasoning capability by aligning the model toward correct intermediate reasoning steps; and (2) \textbf{Step-wise Self-Correction stage}. We hypothesize that modeling preference signals at the step level will provide a stronger foundation for subsequent step-wise self-correction, enabling more effective generation of corrected steps once an incorrect step is identified.

\subsection{Initialization Stage}

In the first stage, we employ a reinforcement learning objective to initialize the model as a initialization training phase. Specifically, it aims to maximize the likelihood of a correct \textsc{solution step} $s_k^+$ while minimizing that of the incorrect \textsc{solution step} $s_k^-$. The optimization objective is:


\begin{align*}
\mathcal{L}_{\text{Pre}}(\theta)
=
- \mathbb{E}_{(x,\{s_i\}_{i=1}^{k-1}, s_k^+, s_k^-)\sim\mathcal{D}}
\Biggl[
\log \sigma \Bigl( \\[-2pt] 
\beta \bigl(
\log \pi_\theta(s_k^+ \mid x,\{s_i\}_{i=1}^{k-1})
-\\[-2pt]
\log \pi_\theta(s_k^- \mid x,\{s_i\}_{i=1}^{k-1})
\bigr)
\Bigr)
\Biggr].
\end{align*}

where $\pi_\theta$ denotes the policy model being optimized, $\beta$ controls the strength of the preference regularization, and $\mathcal{D}$ is a dataset of step-level preference pairs. This stage serves as a initialization phase that establishes a strong foundation for subsequent step-wise self-correction optimization. In this work, we adopt the step-level preference optimization framework of \citet{lai2024step} to implement this initialization stage.

\subsection{Step-wise Self-Correction}

After equipping the model with RL-based step-level supervision signals, we further enable explicit self-correction capability by supervising the model to recognize and correct the erroneous reasoning steps. Formally, given an incorrect reasoning step $s_k^-$ following a correct trajectory $\{s_1, \ldots, s_{k-1}\}$, we construct a preference objective that favors a self-corrected continuation $c_k^+$ over the subsequent incorrect step $s_{k+1}^-$ that would be generated if the error remains unaddressed. The resulting loss:
\begin{align*}
\mathcal{L_{SC}}(\theta)
&= -\mathbb{E}_{
(x,\{s_i\}_{i=1}^{k-1}, s_k^-, c_k^+, s_{k+1}^-)\sim\mathcal{D_{SC}}
}
\Biggl[
\log\sigma\\
&\qquad\Bigl(
\beta\log\frac{\pi_\theta(c_k^+ \mid x,\{s_i\}_{i=1}^{k-1}, s_k^-)}
{\pi_{\text{ref}}(c_k^+ \mid x,\{s_i\}_{i=1}^{k-1}, s_k^-)}
\\&\qquad-
\beta\log\frac{\pi_\theta(s_{k+1}^- \mid x,\{s_i\}_{i=1}^{k-1}, s_k^-)}
{\pi_{\text{ref}}(s_{k+1}^- \mid x,\{s_i\}_{i=1}^{k-1}, s_k^-)}
\Bigr)
\Biggr].
\end{align*}

Here, $\pi_{\text{ref}}$ denotes a frozen baseline model, $s_k^-$ denotes an incorrect reasoning step, while $s_{k+1}^-$ represents the subsequent continuation produced when the error is not detected. In contrast, $c_k^+$ corresponds to a self-corrected reasoning step that explicitly addresses the detected error. We investigate two variants of self-correction supervision that differ in how the corrected step $c_k^+$ is constructed.

\subsubsection{Self-Fix Step-DPO (SFS-DPO)}

Under the SFS-DPO setting, the self-corrected step is constructed by explicitly detecting and fixing an erroneous reasoning step. Formally, we define:
\[
c_k^+ = \{ d_{k-1}, s_k^+ \},
\]
where $d_{k-1}$ is an explicit error-detection signal that flags the flaw in the incorrect step $s_k^-$, and $s_k^+$ is the corrected reasoning step that resolves the detected error. This formulation enables teacher-free self-correction by relying solely on model-generated detection and correction signals.

\subsubsection{Self-Fix Step-DPO with Reasoning (SFS-DPO-R)}

SFS-DPO-R further incorporates external teacher supervision to provide richer corrective signals. Specifically, we leverage a strong teacher model to generate an explicit explanation of why the previous step is incorrect. The corrected step is then defined as:
\[
c_k^+ = \{ d_{k-1}, r_{k-1}, s_k^+ \},
\]
where $r_{k-1}$ denotes a teacher-generated rationale explaining the error in $s_k^-$. By augmenting the correction with high-quality explanatory reasoning, SFS-DPO-R offers stronger supervision at the cost of additional teacher dependence.

We evaluate both SFS-DPO and SFS-DPO-R to contrast a fully teacher-free setting with a teacher-assisted variant, thereby quantifying the trade-off between quality and reliance on external models.

\section{Main Experiments}
\subsection{Experimental Setup}
\paragraph{Baselines.}

To comprehensively evaluate the effectiveness of our proposed methods, we conduct experiments on seven widely used open-source LLMs, following the original Step-DPO experimental setup. Specifically, we use three backbone LLMs from Step-DPO: DeepSeekMath-7B \cite{shao2024deepseekmath}, Qwen2-7B, and Qwen2-7B-Instruct \cite{bai2023qwen}. DeepSeekMath-7B and Qwen2-7B are further fine-tuned on the MetaMath \cite{yu2024metamath} and MMIQC \cite{liu2025augmenting} datasets, yielding DeepSeekMath-7B-SFT and Qwen2-7B-SFT. In addition, we include three stronger recent instruction-tuned models, Llama-3.1-8B-Instruct \cite{grattafiori2024llama}, Qwen3-8B \cite{yang2025qwen3} and Qwen2.5-14B-Instruct \cite{Yang2024Qwen25TR}, as well as the recent math-specialized model Qwen2.5-Math-7B-Instruct \cite{yang2024qwen2}, to further assess the scalability and robustness of our methods across both general-purpose and math-oriented reasoning LLMs.

\definecolor{softdarkgreen}{rgb}{0.0,0.45,0.25}

\begin{table*}[ht]
\centering
\small
\renewcommand{\arraystretch}{1}
\setlength{\tabcolsep}{10pt}
\begin{tabular}{l|cc|cc|c}
\toprule
& \multicolumn{2}{c|}{\textbf{In-Domain}} & \multicolumn{2}{c|}{\textbf{Out-of-Domain}} & \\
\textbf{Model} & \textbf{MATH} & \textbf{GSM8K} & \textbf{GK2023} & \textbf{OCW} & \textbf{Avg.} \\
\midrule
DeepSeekMath-7B-SFT & 51.7 & 86.8 & 38.2 & 19.1 & 49.0 \\
\quad + Step-DPO & 51.8$_{\textcolor{softdarkgreen}{(+0.1)}}$ & 86.7$_{\textcolor{red}{(-0.1)}}$ & 43.4$_{\textcolor{softdarkgreen}{(+5.2)}}$ & 18.0$_{\textcolor{red}{(-1.1)}}$ & 50.0$_{\textcolor{softdarkgreen}{(+1.0)}}$ \\
\rowcolor{blue!10} \quad + SFS-DPO & \textbf{52.2}$_{\textcolor{softdarkgreen}{(+0.5)}}^{*}$ & 87.6$_{\textcolor{softdarkgreen}{(+0.8)}}^{*}$ & \textbf{43.9}$_{\textcolor{softdarkgreen}{(+5.7)}}$ & 23.2$_{\textcolor{softdarkgreen}{(+4.1)}}$ & 51.7$_{\textcolor{softdarkgreen}{(+2.7)}}$ \\
\rowcolor{blue!10} \quad + SFS-DPO-R & \textbf{52.2}$_{\textcolor{softdarkgreen}{(+0.5)}}^{*}$ & \textbf{88.0}$_{\textcolor{softdarkgreen}{(+1.2)}}^{*}$ & \textbf{43.9}$_{\textcolor{softdarkgreen}{(+5.7)}}$ & \textbf{25.0}$_{\textcolor{softdarkgreen}{(+5.9)}}$ & \textbf{52.3}$_{\textcolor{softdarkgreen}{(+3.3)}}$ \\
\midrule
Qwen2-7B-SFT & 53.9 & 87.6 & 46.2 & 15.8 & 50.9 \\
\quad + Step-DPO & 55.3$_{\textcolor{softdarkgreen}{(+1.4)}}$ & 87.6$_{\textcolor{softdarkgreen}{(+0.0)}}$ & 45.7$_{\textcolor{red}{(-0.5)}}$ & 15.8$_{\textcolor{softdarkgreen}{(+0.0)}}$ & 51.1$_{\textcolor{softdarkgreen}{(+0.2)}}$ \\
\rowcolor{blue!10} \quad + SFS-DPO & \textbf{55.6}$_{\textcolor{softdarkgreen}{(+1.7)}}^{*}$ & 87.9$_{\textcolor{softdarkgreen}{(+0.3)}}$ & 46.0$_{\textcolor{red}{(-0.2)}}$ & \textbf{23.9}$_{\textcolor{softdarkgreen}{(+8.1)}}$ & \textbf{53.4}$_{\textcolor{softdarkgreen}{(+2.5)}}$ \\
\rowcolor{blue!10} \quad + SFS-DPO-R & 55.4$_{\textcolor{softdarkgreen}{(+1.5)}}$ & \textbf{88.0}$_{\textcolor{softdarkgreen}{(+0.4)}}^{*}$ & \textbf{46.2}$_{\textcolor{softdarkgreen}{(+0.0)}}$ & 22.8$_{\textcolor{softdarkgreen}{(+7.0)}}$ & 53.1$_{\textcolor{softdarkgreen}{(+2.2)}}$ \\
\midrule
Qwen2-7B-Instruct & 55.7 & 85.0 & 39.7 & 20.2 & 50.2 \\
\quad + Step-DPO & 57.1$_{\textcolor{softdarkgreen}{(+1.4)}}$ & 86.2$_{\textcolor{softdarkgreen}{(+1.2)}}$ & 42.9$_{\textcolor{softdarkgreen}{(+3.2)}}$ & 21.3$_{\textcolor{softdarkgreen}{(+1.1)}}$ & 51.9$_{\textcolor{softdarkgreen}{(+1.7)}}$ \\
\rowcolor{blue!10} \quad + SFS-DPO & 58.6$_{\textcolor{softdarkgreen}{(+2.9)}}^{*}$ & \textbf{86.6}$_{\textcolor{softdarkgreen}{(+1.6)}}^{*}$ & \textbf{46.0}$_{\textcolor{softdarkgreen}{(+6.3)}}$ & 20.6$_{\textcolor{softdarkgreen}{(+0.4)}}$ & \textbf{53.0}$_{\textcolor{softdarkgreen}{(+2.8)}}$ \\
\rowcolor{blue!10} \quad + SFS-DPO-R & \textbf{59.1}$_{\textcolor{softdarkgreen}{(+3.4)}}^{*}$ & 86.0$_{\textcolor{softdarkgreen}{(+1.0)}}$ & 44.7$_{\textcolor{softdarkgreen}{(+5.0)}}$ & \textbf{22.1}$_{\textcolor{softdarkgreen}{(+1.9)}}$ & \textbf{53.0}$_{\textcolor{softdarkgreen}{(+2.8)}}$ \\
\midrule
Qwen2.5-Math-7B-Instruct & 83.6 & 95.2 & 57.4 & 23.9 & 65.0 \\
\quad + Step-DPO & 84.6$_{\textcolor{softdarkgreen}{(+1.0)}}$ & 95.8$_{\textcolor{softdarkgreen}{(+0.6)}}$ & 67.0$_{\textcolor{softdarkgreen}{(+9.6)}}$ & 31.6$_{\textcolor{softdarkgreen}{(+7.7)}}$ & 69.8$_{\textcolor{softdarkgreen}{(+4.8)}}$ \\
\rowcolor{blue!10} \quad + SFS-DPO & 84.7$_{\textcolor{softdarkgreen}{(+1.1)}}$ & 95.8$_{\textcolor{softdarkgreen}{(+0.6)}}$ & \textbf{68.3}$_{\textcolor{softdarkgreen}{(+10.9)}}$ & 32.0$_{\textcolor{softdarkgreen}{(+8.1)}}$ & 70.2$_{\textcolor{softdarkgreen}{(+5.2)}}$ \\
\rowcolor{blue!10} \quad + SFS-DPO-R & \textbf{85.0}$_{\textcolor{softdarkgreen}{(+1.4)}}^{*}$ & \textbf{96.0}$_{\textcolor{softdarkgreen}{(+0.8)}}$ & 67.8$_{\textcolor{softdarkgreen}{(+10.4)}}$ & \textbf{32.7}$_{\textcolor{softdarkgreen}{(+8.8)}}$ & \textbf{70.4}$_{\textcolor{softdarkgreen}{(+5.4)}}$ \\
\midrule
Qwen3-8B & 72.9 & 92.8 & 58.4 & 31.3 & 63.9 \\
\quad + Step-DPO & 73.4$_{\textcolor{softdarkgreen}{(+0.5)}}$ & 92.8$_{\textcolor{softdarkgreen}{(+0.0)}}$ & 58.2$_{\textcolor{red}{(-0.2)}}$ & 31.3$_{\textcolor{softdarkgreen}{(+0.0)}}$ & 63.9$_{\textcolor{softdarkgreen}{(+0.0)}}$ \\
\rowcolor{blue!10} \quad + SFS-DPO & 73.5$_{\textcolor{softdarkgreen}{(+0.6)}}$ & 93.1$_{\textcolor{softdarkgreen}{(+0.3)}}$ & 58.4$_{\textcolor{softdarkgreen}{(+0.0)}}$ & 33.1$_{\textcolor{softdarkgreen}{(+1.8)}}$ & 64.5$_{\textcolor{softdarkgreen}{(+0.6)}}$ \\
\rowcolor{blue!10} \quad + SFS-DPO-R & \textbf{73.7}$_{\textcolor{softdarkgreen}{(+0.8)}}^{*}$ & \textbf{93.9}$_{\textcolor{softdarkgreen}{(+1.1)}}^{*}$ & \textbf{59.0}$_{\textcolor{softdarkgreen}{(+0.6)}}$ & \textbf{34.2}$_{\textcolor{softdarkgreen}{(+2.9)}}$ & \textbf{65.2}$_{\textcolor{softdarkgreen}{(+1.3)}}$ \\
\midrule
Llama-3.1-8B-Instruct & 50.4 & 86.2 & 38.4 & 25.0 & 50.0 \\
\quad + Step-DPO & \textbf{51.8}$_{\textcolor{softdarkgreen}{(+1.4)}}$ & 86.4$_{\textcolor{softdarkgreen}{(+0.2)}}$ & 41.8$_{\textcolor{softdarkgreen}{(+3.4)}}$ & 27.6$_{\textcolor{softdarkgreen}{(+2.6)}}$ & 51.9$_{\textcolor{softdarkgreen}{(+1.9)}}$ \\
\rowcolor{blue!10} \quad + SFS-DPO & 51.0$_{\textcolor{softdarkgreen}{(+0.6)}}$ & 86.7$_{\textcolor{softdarkgreen}{(+0.5)}}$ & 42.1$_{\textcolor{softdarkgreen}{(+3.7)}}$ & \textbf{28.3}$_{\textcolor{softdarkgreen}{(+3.3)}}$ & 52.0$_{\textcolor{softdarkgreen}{(+2.0)}}$ \\
\rowcolor{blue!10} \quad + SFS-DPO-R & 51.7$_{\textcolor{softdarkgreen}{(+1.3)}}$ & \textbf{87.1}$_{\textcolor{softdarkgreen}{(+0.9)}}^{*}$ & \textbf{42.6}$_{\textcolor{softdarkgreen}{(+4.2)}}$ & 27.9$_{\textcolor{softdarkgreen}{(+2.9)}}$ & \textbf{52.3}$_{\textcolor{softdarkgreen}{(+2.3)}}$ \\
\midrule
Qwen2.5-14B-Instruct & 78.8 & 93.8 & 65.5 & 37.5 & 68.9 \\
\quad + Step-DPO & 79.3$_{\textcolor{softdarkgreen}{(+0.5)}}$ & 94.1$_{\textcolor{softdarkgreen}{(+0.3)}}$ & 65.7$_{\textcolor{softdarkgreen}{(+0.2)}}$ & 36.0$_{\textcolor{red}{(-1.5)}}$ & 68.8$_{\textcolor{red}{(-0.1)}}$ \\
\rowcolor{blue!10} \quad + SFS-DPO & 79.2$_{\textcolor{softdarkgreen}{(+0.4)}}$ & \textbf{94.5}$_{\textcolor{softdarkgreen}{(+0.7)}}^{*}$ & 65.7$_{\textcolor{softdarkgreen}{(+0.2)}}$ & \textbf{38.2}$_{\textcolor{softdarkgreen}{(+0.7)}}$ & 69.4$_{\textcolor{softdarkgreen}{(+0.5)}}$ \\
\rowcolor{blue!10} \quad + SFS-DPO-R & \textbf{79.4}$_{\textcolor{softdarkgreen}{(+0.6)}}$ & \textbf{94.5}$_{\textcolor{softdarkgreen}{(+0.7)}}^{*}$ & \textbf{67.8}$_{\textcolor{softdarkgreen}{(+2.3)}}$ & 37.7$_{\textcolor{softdarkgreen}{(+0.2)}}$ & \textbf{69.9}$_{\textcolor{softdarkgreen}{(+1.0)}}$ \\
\bottomrule
\end{tabular}
\caption{Percentage accuracy on in-domain (MATH, GSM8K) and out-of-domain (GK2023, OCW) benchmarks with greedy decoding. Numbers in parentheses denote absolute accuracy change relative to the corresponding base backbone. For in-domain benchmarks, * denotes statistically significant improvements over Step-DPO according to one-sided McNemar's test (p < 0.05).}
\label{tab:main-experiments}
\end{table*}

\paragraph{Dataset Construction \& Training Setup.} 

To construct preference learning data for the step-wise self-correction stage, we first collect samples from the original 10K dataset of \citet{lai2024step}. For each sample, we then append the rejected reasoning step to the initial reasoning to form a new reasoning prefix with an erroneous reasoning step, and use the subsequent rejected reasoning step as the rejected sample for our dataset. 
Following \citet{pan-etal-2025-lemma}, to construct the chosen step, we concatenate the correct reasoning step with a self-correction signal. The same set of signal phrases, listed in \Cref{fig:sfsdpo_construction}, is then used to automatically identify self-correction behavior in generated responses.
Through this process, we obtain a resource-free SFS-DPO dataset containing 8{,}416 samples. We further leverage GPT-4o \cite{hurst2024gpt} to generate explicit explanations for incorrect reasoning steps and insert them between the self-correction signal and the correct reasoning step, resulting in the SFS-DPO-R dataset. Further details are provided in \Cref{fig:sfsdpoexample}.

In the initialization stage, to ensure a fair comparison, we use the same 10K-sample training dataset from Step-DPO \cite{lai2024step}, which is collected from the training set of GSM8K \cite{cobbe2021trainingverifierssolvemath} and MATH \cite{hendrycks2021measuring}, as the initial source of step-level reasoning supervision. 
All models are first trained for three epochs in the initialization stage with batch size 4. We then train SFS-DPO and SFS-DPO-R for four epochs in the self-correction stage using batch size 8. All training stages use the AdamW optimizer with a warmup ratio of 0.02 and learning rate of $5\times10^{-7}$.


\paragraph{Benchmarks.}
To evaluate our method, we consider both in-domain and out-of-domain (OOD) benchmarks. In-domain performance is measured on \textbf{GSM8K} \cite{cobbe2021trainingverifierssolvemath} and \textbf{MATH} \cite{hendrycks2021measuring}, with 1,319 and 5,000 test questions, respectively. To assess generalization, we report results on \textbf{GaoKao2023} (GK2023) \citep{gk2023} comprising 385 competition-level math problems from the 2023 Chinese university entrance exam, and \textbf{OCWCourses} (OCW) \citep{ocw} containing 272 undergraduate-level STEM problems requiring multi-step reasoning. All benchmarks are evaluated using answer accuracy. We define the self-correction rate metric as the fraction of model-generated solutions that are correct, conditioned on the model deciding to perform self-correction. Following \citet{ma2025s}, we further report Error Recall, the fraction of incorrect reasoning steps that the model flags via a self-correction signal.



\subsection{In-Domain Results}

Our main experimental results in \Cref{tab:main-experiments} demonstrate that explicitly modeling step-level self-correction yields stronger gains than prior step-wise preference learning. Across seven backbones and two in-domain datasets, Step-DPO provides limited and inconsistent improvements. While SFS-DPO and SFS-DPO-R improve over the base models in all settings, Step-DPO mostly gives gains below 1\%, with performance plateauing or slightly degrading on GSM8K for Qwen2-7B-SFT and DeepSeekMath-7B-SFT. In contrast, SFS-DPO and SFS-DPO-R learn explicit self-correction and achieve larger gains than Step-DPO, outperforming it in 11/14 and 12/14 settings, respectively. This suggests that self-correction training improves reasoning performance while teaching models when and how to revise their reasoning.


On the model level, our method yields average improvements of 0.83/0.97\% on the math-specialized backbones, and even larger gains of 0.95/1.23\% on the instruction-tuned backbones. Qwen2-7B-Instruct achieves the most noticeable gains, with a 3.4\% improvement on MATH under SFS-DPO-R. Improvements are smaller on recent backbones, namely Qwen2.5-14B-Instruct, Qwen3-8B and Llama-3.1-8B-Instruct. One likely factor is that these backbones already self-correct more often before training, leaving less room to introduce new correction behavior; further analysis is provided in \Cref{analysis}. Overall, these results highlight that the primary advantage of our method lies not in finer-grained preference learning alone, but in optimizing preferences over corrected trajectories.



Averaged across seven backbones, SFS-DPO improves accuracy by 1.11\% on MATH and 0.69\% on GSM8K, while SFS-DPO-R further increases the gains to 1.36\% and 0.87\%, respectively. The consistent advantage of SFS-DPO-R over SFS-DPO indicates that explicit explanatory reasoning about errors provides additional supervision beyond correction alone. By exposing the model to why a reasoning step is incorrect, SFS-DPO-R encourages more accurate error localization and more targeted repairs, leading to improved downstream accuracy.

\subsection{Out-Of-Domain Results (OOD)}

To assess robustness under distribution shift, we evaluate on the competition-level GK2023 benchmark and undergraduate OCWCourses (OCW), as shown in \Cref{tab:main-experiments}. Step-DPO shows inconsistent OOD behavior, including degradation on Qwen2-7B-SFT for GK2023 and on the larger Qwen2.5-14B-Instruct for OCW, while SFS-DPO and SFS-DPO-R consistently maintain or improve accuracy across all backbones and datasets. 
Notably, Qwen2-7B-Instruct achieves a substantial 6.3\% improvement on GK2023 under SFS-DPO, while Qwen2.5-Math-7B-Instruct achieves the largest overall OOD gains, reaching 10.9\% on GK2023 and 8.8\% on OCW.
These results indicate that explicit step-level error detection and correction yields correction strategies that generalize beyond the training distribution, while the small gap between SFS-DPO and SFS-DPO-R suggests that self-generated correction signals already capture much of the transferable structure needed for OOD reasoning.

\begin{figure*}
    \centering
    \includegraphics[width=\linewidth]{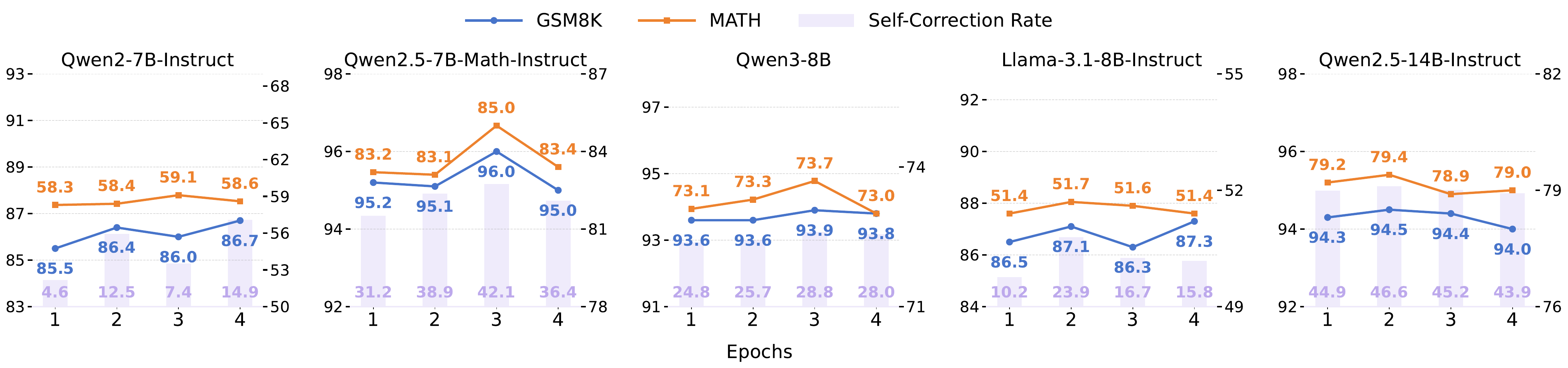}
    \caption{Self-correction rate of SFS-DPO-R under five Instruct LLMs. Self-correction rate shows a positive correlation with task accuracy}
    \label{fig:placeholder}
\end{figure*}

\begin{figure}
    \centering
    \includegraphics[width=\linewidth]{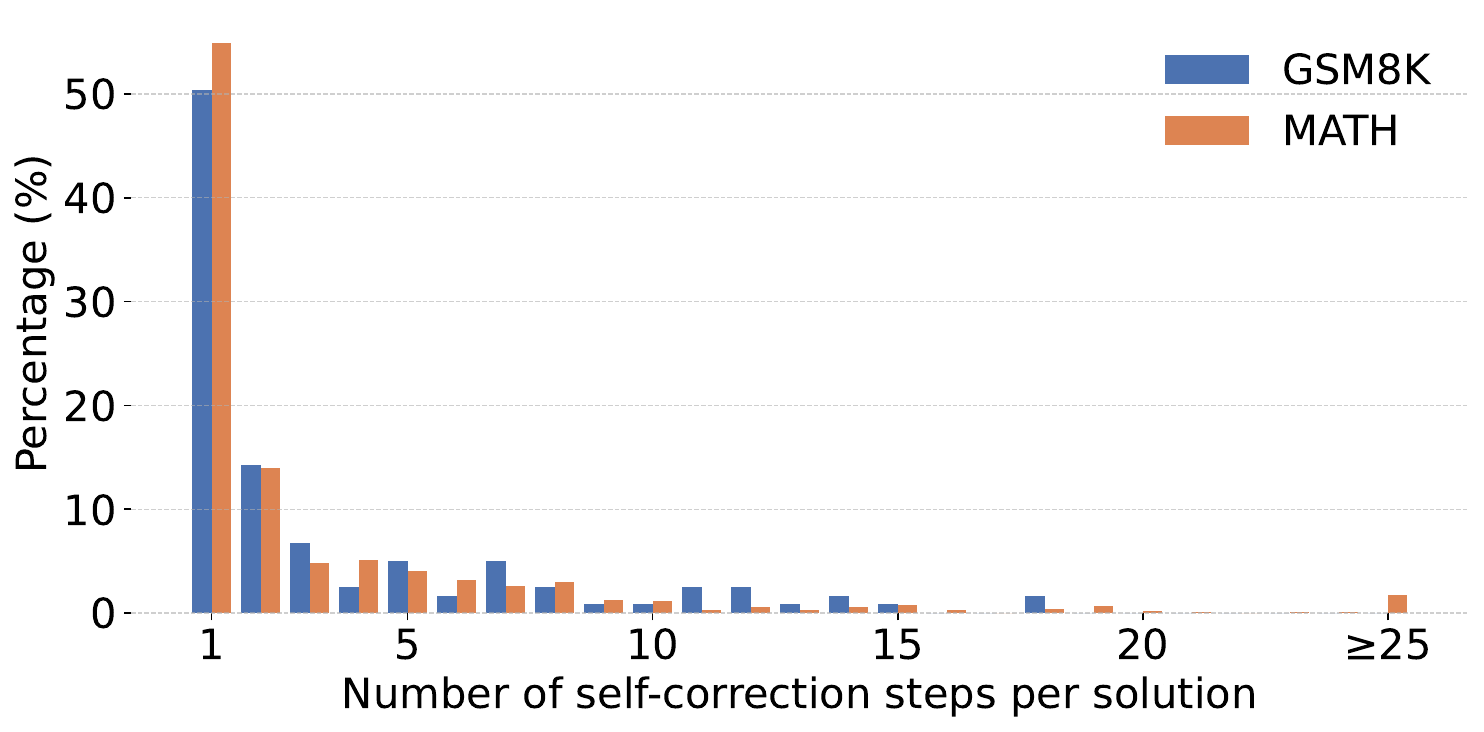}
    \caption{Distribution of numbers of self-correction steps per solution under SFS-DPO-R, averaged over six backbone models.}
    \label{SC_distribution}
\end{figure}

\section{Discussions}


\subsection{Comparison to Self-Correction Baselines}

\begin{table}[H]
\centering
{\resizebox{\columnwidth}{!}{
\renewcommand{\arraystretch}{1} 
\setlength{\tabcolsep}{4pt} 
\begin{tabular}{l|c|c|c|c}
\toprule
    \textbf{Model} & \textbf{MATH} & \textbf{GSM8K} & \textbf{SC Rate} & \textbf{Error Recall}\\
\midrule
    LEMMA & 48.5 &  83.3 & 45.3 & 49.9\\
    S$^2$R  & 48.7 &  84.4 & 46.5 & 54.6 \\
    \rowcolor{blue!10} SFS-DPO & 51.0 & 86.7 & 28.8 & 33.2 \\
    \rowcolor{blue!10} SFS-DPO-R & \textbf{51.7} & \textbf{87.1}  & 23.9  & 35.9 \\
\bottomrule
\end{tabular}}}
\caption{Comparison of prior self-correction methods and ours with Llama-3.1-8B-Instruct as the backbone. }
\label{tab:sc-with-baseline}
\end{table}


We compare our framework with prior self-correction supervised fine-tuning (SFT) baselines. Due to the availability of comparable baselines, we conduct all experiments using Llama-3.1-8B-Instruct as the shared backbone. As shown in \Cref{tab:sc-with-baseline}, SFS-DPO and SFS-DPO-R consistently outperform existing baselines on both MATH and GSM8K while exhibiting lower self-correction rates and lower Error Recall, indicating that neither correcting more often nor detecting more errors translates into better performance. This may be due to SFT-based methods suffering from distribution shift and behavior collapse \cite{kumar2024training}, which biases the model toward excessive self-correction: it follows correction templates rather than selectively revising genuine errors. Further analysis of these behavioral metrics is discussed in \Cref{sc_rate_anal}.
\subsection{The Role of Initialization Stage}

\begin{table}[H]
\centering
\small{\resizebox{0.99\columnwidth}{!}{
\renewcommand{\arraystretch}{1.15}
\setlength{\tabcolsep}{6pt}
\begin{tabular}{c|l|c|c}
\toprule
 & \textbf{Initialization} & \textbf{MATH} & \textbf{GSM8K} \\
\midrule

\multirow{4}{*}{\rotatebox{90}{\textbf{SFT}}}


& Step RL (ours)
& 55.6 
& 87.9  \\

& No init.
& 53.1 \textcolor{red}{\scriptsize(-2.5)}
& 87.3 \textcolor{red}{\scriptsize(-0.6)} \\

& No init. + Joint Training
& 53.3 \textcolor{red}{\scriptsize(-2.2)}
& 84.1 \textcolor{red}{\scriptsize(-3.8)} \\

& RL   
& 53.3 \textcolor{red}{\scriptsize(-2.2)}  
& 83.6 \textcolor{red}{\scriptsize(-4.3)} \\

\midrule

\multirow{4}{*}{\rotatebox{90}{\textbf{Instruct}}}

& Step RL (ours)          
& 58.6 & 86.6 \\

& No init.     
& 53.1 \textcolor{red}{\scriptsize(-4.1)}  
& 85.6 \textcolor{red}{\scriptsize(-1.0)} \\

& No init. + Joint Training
& 56.9 \textcolor{red}{\scriptsize(-1.7)}  
& 86.4 \textcolor{red}{\scriptsize(-0.2)} \\

& RL    
& 55.7 \textcolor{red}{\scriptsize(-0.9)}  
& 86.4 \textcolor{red}{\scriptsize(-0.2)} \\

\bottomrule
\end{tabular}}}
\caption{Comparison of training strategies using Qwen2-7B models. Results in percentage are reported relative to our SFS-DPO baseline.}
\label{tab:step-dpo-ablation}
\end{table}

\Cref{tab:step-dpo-ablation} compares our step-level initialization with three alternatives: no initialization, joint training of reasoning and self-correction ability, and standard RL initialization. Removing the initialization stage leads to clear performance degradation, showing that directly optimizing self-correction signals without a strong reasoning foundation is insufficient. In addition, joint training is also detrimental, suggesting that step-wise reasoning and self-correction are better learned sequentially rather than as a single mixed objective. Standard RL initialization further underperforms Step RL, indicating that step-level preference optimization provides a stronger foundation for later self-correction training. Furthermore, the degradation is especially pronounced on the more challenging MATH benchmark and in the SFT setting. Overall, these results highlight the importance of establishing strong step-level reasoning before optimizing self-correction behavior.

\begin{figure}
    \centering
    \includegraphics[width=\linewidth]{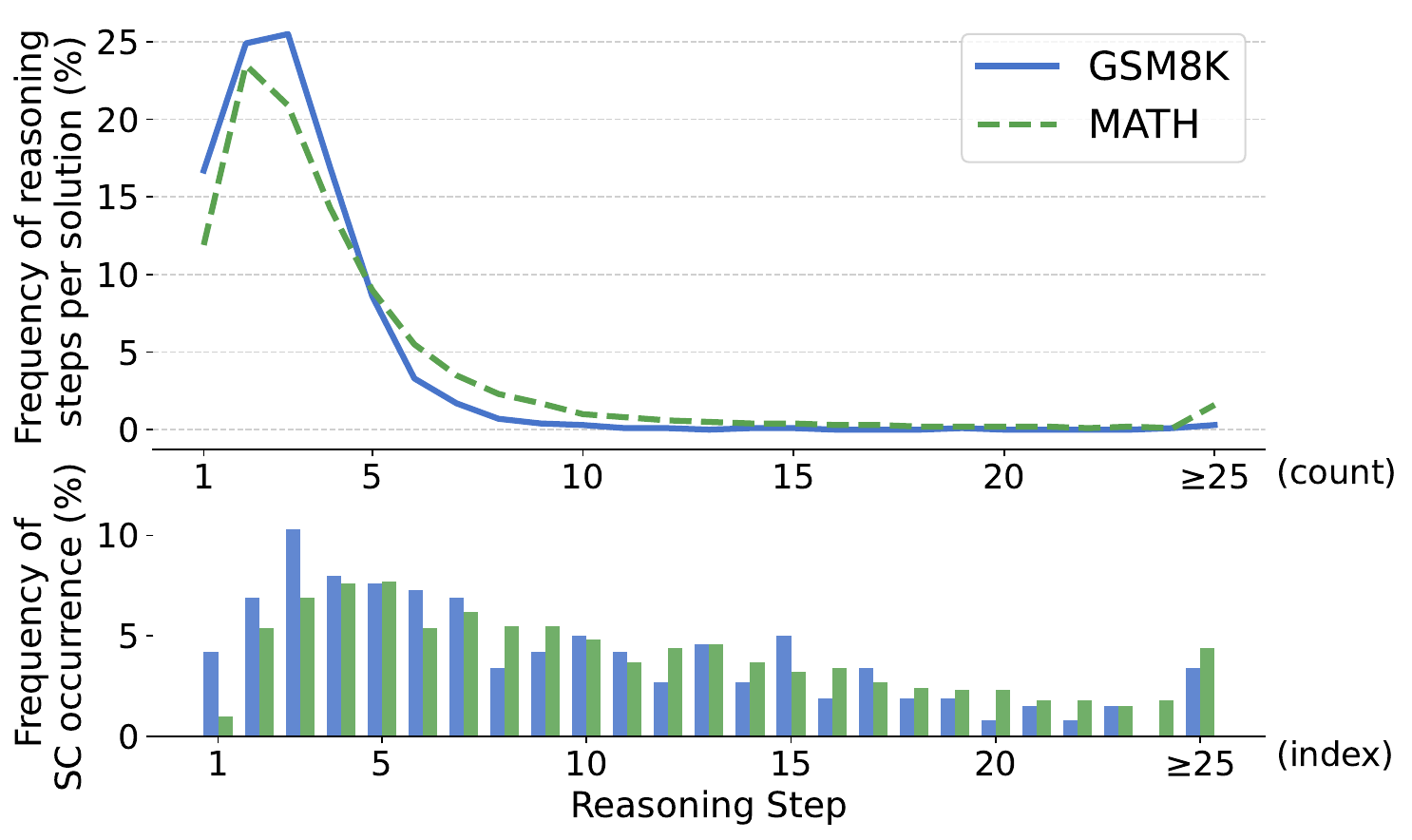}
    \caption{Distribution of self-correction behavior under SFS-DPO-R, averaged over six backbone models.}
    \label{SC_frequency}
\end{figure}

\begin{table*}[t]
\centering
\small
\renewcommand{\arraystretch}{1.15}
\setlength{\tabcolsep}{6pt}

\begin{tabular}{lp{13cm}}
\toprule
Problem & A custodian has to clean a school with 80 classrooms. They have 5 days to get it done. It takes them 15 minutes per classroom. If they work an 8 hour day, what percentage of their day, on average, is spent cleaning classrooms?
\\
\midrule
Base Model
&  
... Step 3: Now, we need to find out what percentage of the day is spent 
cleaning classrooms. To do this, we divide the time spent cleaning classrooms 
(1200 minutes) by the total time in a day (480 minutes) and multiply by 100 
to get the percentage. So the percentage is $(1200/480) \times 100 = 250\%$.

\textcolor{red}{However, this result is not possible because a percentage cannot be more than 100\%. 
There must be a mistake in our calculation. Let's retrace our steps. \newline
Step 4: We found that it takes 1200 minutes to clean all the classrooms 
and there are 480 minutes in a day. So the percentage of the day spent cleaning 
classrooms should be $(1200/480) \times 100 = 250\%$. 
\newline
<< Repeated block >> \newline
<< Repeated block >> \newline
The mistake is in the calculation of the percentage of the day spent 
cleaning classrooms. The percentage of the day spent cleaning classrooms 
should be the time spent cleaning classrooms divided by the total time in a day, which is $(1200/480) \times 100 = 250$}
 \\
\midrule
\textbf{SFS-DPO-R} &

... Step 2: The conversion of hours to minutes is also correct. It is indeed 20 hours.

\textcolor{red}{Step 3: The calculation of the percentage of the day spent cleaning is incorrect. We should have divided the time spent cleaning (20 hours) by the total work hours (8 hours) and then multiplied by 100 to get the percentage. The correct calculation is $(20 / 8) * 100 = 250\%$.}

\textcolor{orange}{The mistake was in the last step. We should not have multiplied by 100 after dividing. The correct calculation should be $(20 / 8) * 100 = 250\%$. But this doesn't make sense because the percentage cannot be more than $100\%$. The mistake was in the last step. We should not have multiplied by 100 after dividing. The mistake was in the calculation of the total work hours. The custodian works an 8-hour day, but they have 5 days to clean the classrooms. So the total work hours should be 8 hours/day * 5 days = 40 hours.}

\textcolor{Green}{Now let's correct the calculation.
Step 3: The correct calculation of the percentage of the day spent cleaning is $(20 / 40) * 100 = 50\%$.
So, on average, the custodian spends $50\%$ of their day cleaning classrooms.
The answer is: 50}\\
 \bottomrule
\end{tabular}
\caption{Qualitative reasoning between Qwen2-7B-SFT and Qwen2-7B-SFT-SFS-DPO-R. 
}
\label{tab:sfsdpo_example}
\end{table*}

\subsection{Self-Correction Rate}
\label{sc_rate_anal}


We study the role of the self-correction (SC) rate in model performance. As shown in \Cref{fig:placeholder}, the SC rate exhibits a positive correlation with overall performance, indicating that the self-correction ability of our framework plays an important role in improving reasoning accuracy. However, a high SC rate does not necessarily translate into better accuracy. Across MATH and GSM8K, some self-correction methods achieve high SC rates while still underperforming in final task accuracy (\Cref{tab:sc-with-baseline}), suggesting that excessive self-correction may reflect forced correction behavior rather than effective reasoning improvement. In contrast, our methods achieve stronger overall performance despite lower SC rates, indicating that our framework learns to apply self-correction more selectively and effectively. Rather than maximizing correction frequency, our method encourages the model to self-correct primarily on more challenging examples where the original reasoning is likely to fail. This suggests that effective self-correction requires not only knowing how to self-correct, but also knowing when not to self-correct. Overall, these findings indicate that SC rate alone is insufficient as a standalone indicator of self-correction quality. What ultimately matters is whether self-correction behavior is positively aligned with task performance, such that revisions reinforce correct reasoning instead of introducing unnecessary or spurious changes.

\subsection{Self-Correction Behavior}

\Cref{SC_distribution} and \Cref{SC_frequency} present a detailed analysis of the self-correction behavior exhibited by the models on GSM8K and MATH. \Cref{SC_distribution} analyzes the distribution of the total number of self-correction steps in solutions, conditioned on the presence of the self-correction signal. It shows that most solutions contain only a few self-correction steps, with single-correction cases being most common and the frequency decreasing as corrections increase. Cases with more than three corrections are relatively uncommon, suggesting that the model does not rely on repeated or forced revisions. \Cref{SC_frequency} further shows that self-correction occurrences broadly follow the distribution of reasoning lengths rather than concentrating at specific positions, indicating no strong positional bias in self-correction behavior toward specific reasoning stages. Overall, these results suggest that SFS-DPO-R encourages stable and targeted self-correction behavior during multi-step reasoning, providing a foundation for learning when to self-correct effectively and selectively. 

\subsection{Case Study}

As shown in \Cref{tab:sfsdpo_example}, the base Qwen2-7B-SFT model makes an error in its calculation and repeatedly generates incorrect reasoning without effectively resolving the mistake. In contrast, SFS-DPO-R enables the model to detect the erroneous reasoning step, localize the calculation error in the middle of the solution, and explicitly reason about the source of the mistake. Additional qualitative examples are provided in the \Cref{sec:appendix}.

\section{Conclusion}
In this work, we propose SFS-DPO, a two-stage framework that enables small LLMs to explicitly detect and correct erroneous reasoning steps during inference. The framework builds on step-level preference optimization, with an RL-based initialization to strengthen step-wise reasoning, followed by targeted training for self-correction. We further introduce SFS-DPO-R, a teacher-assisted variant that incorporates explanatory rationales to provide stronger corrective signals. Comprehensive experiments across multiple model backbones demonstrate consistent improvements over prior methods on both in-domain and out-of-distribution benchmarks. These results highlight that explicitly modeling how errors are identified and fixed, rather than merely preferring better continuations, is critical for robust math reasoning, positioning step-wise self-correction as a promising direction for improving the reliability of large language models.


\section*{Limitations}

While our framework represents an advancement in self-correction capabilities during inference of LLMs, several limitations persist. Firstly, SFS-DPO-R relies on stronger models' generated rationales, introducing additional teacher dependence and potential propagation of teacher biases or errors. In addition, our evaluation focuses on mathematical reasoning tasks, where intermediate reasoning steps are naturally well-defined. The generalization of the proposed method to more open-ended settings, such as creative writing, has not been fully explored. Future work will explore scaling this framework to more complex datasets. Finally, we examined moderately sized LLMs, ranging from 7 to 14 billion parameters. Experiments with larger and more capable models could strengthen our claims.

\section*{Acknowledgement}

This research is supported by the RIE2025 Industry Alignment Fund – Industry Collaboration Projects (IAF-ICP)(Award I2301E0026), administered by A*STAR, as well as supported by Alibaba Group and NTU Singapore
through Alibaba-NTU Global e-Sustainability CorpLab (ANGEL). Do Xuan Long is supported by the A*STAR Computing and Information Science (ACIS) Scholarship.

\bibliography{custom}

\newpage
\onecolumn

\appendix
\twocolumn
\appendix

\section{Choice of Statistical Test}

For the main benchmark results, where the evaluation outcomes are binary (correct/incorrect), we adopt McNemar’s test for the significant testing. This test is particularly suitable for paired binary evaluation settings, as it directly compares the prediction differences between two systems on the same evaluation samples. Specifically, McNemar’s test focuses on discordant pairs, i.e., cases where the prediction outcomes differ between our method and the Step-DPO baseline. We use a one-sided exact McNemar’s test under the hypothesis that our method improves over the baseline and report results with $p < 0.05$ as statistically significant. 

We only report McNemar’s test results for the main benchmarks due to their relatively larger evaluation sizes. For smaller OOD benchmarks, we do not report significance results since the limited number of evaluation samples makes McNemar’s test statistically underpowered and less reliable.

\section{Additional Analysis}

\label{analysis}

We present additional analysis of self-correction behavior before and after applying SFS-DPO and SFS-DPO-R in \Cref{tab:sc-sfsdpo}. Before training, several backbone models, including DeepSeekMath-7B-SFT, Qwen2-7B-SFT and Qwen2-7B-Instruct show limited self-correction ability. After training, SFS-DPO and SFS-DPO-R generally improve self-correction behavior on these models. However, a higher self-correction rate does not necessarily imply better reasoning performance, as shown by Llama-3.1-8B-Instruct and  Qwen3-8B, where the self-correction rate decreases after training while the overall reasoning performance improves. This suggests that effective self-correction depends on selective and accurate revision rather than frequency alone.

\begin{figure}[H]
    \centering
    \includegraphics[width=\linewidth]{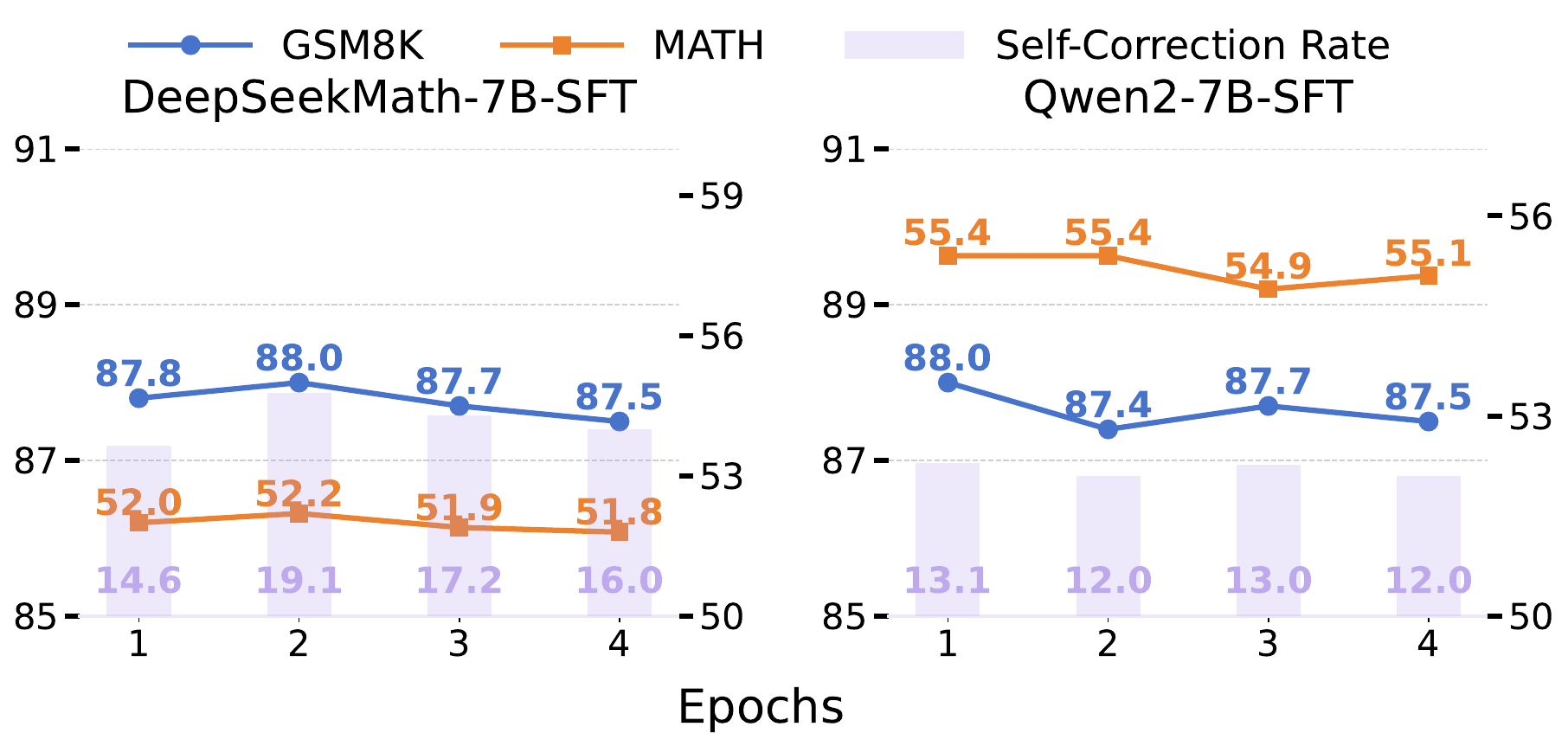}
    \caption{Self-correction rate of SFS-DPO-R under two SFT models. }
    \label{SC_distribution_appendix}
\end{figure}

Figure~\ref{SC_distribution_appendix} further analyzes epoch-wise behavior on the two SFT backbones. For DeepSeekMath-7B-SFT, both self-correction rate and task accuracy peak at epoch 2, suggesting positive alignment between correction behavior and reasoning performance. In contrast, Qwen2-7B-SFT shows relatively stable self-correction rates with only minor accuracy fluctuations, again indicating that correction effectiveness and selectivity matter more than raw correction frequency.

\begin{table}[H]
\centering

\renewcommand{\arraystretch}{1} 
\setlength{\tabcolsep}{4pt} 
\small
\begin{tabular}{l|c|c|c}
\toprule
\textbf{Model} & \textbf{Base} & \textbf{SFS-DPO}  &  \textbf{SFS-DPO-R} \\
\midrule
DeepSeekMath-7B-SFT & 8.1 & 14.2 & 19.1 \\
Qwen2-7B-SFT & 8.0 & 14.2 & 13.1 \\
Qwen2-7B-Instruct & 5.3 & 12.5 & 14.9 \\
Qwen2.5-7B-Math-Instr. & 21.0 & 35.7 & 42.1 \\
Qwen3-8B & 35.8 & 26.9 & 28.8 \\
Llama-3.1-8B-Instruct & 37.0 & 28.8 & 23.9 \\
Qwen-2.5-14B-Instruct & 37.5 & 43.8 & 46.6 \\

\bottomrule
\end{tabular}
\caption{Self-correction rates of SFS-DPO and SFS-DPO-R across different backbone models.}
\label{tab:sc-sfsdpo}
\end{table}

\section{Data Construction Details}

We provide additional details on our data construction process. As shown in \Cref{fig:sfsdpo_construction}, we use a set of transition signals adapted from \citet{pan-etal-2025-lemma} to indicate the beginning of self-correction behavior. We further extend this list with additional explicit phrases, such as ``The previous step is incorrect.'' and ``There is a mistake.'', to encourage clearer error-detection behavior before generating the corrected reasoning step.

\begin{figure*}[htbp]
    \small
    \begin{tcolorbox}[left=3pt,right=3pt,top=3pt,bottom=3pt,title=\textbf{``Self-Fix Step-DPO'' Prompt Construction}]
    
    \textbf{\{Question\}}. Let's think step by step.
    
    \textbf{\{Annotated Incorrect Reasoning Steps\}}

    \textbf{\{Choose one transition phrase below\}}
    \begin{itemize}
        \item But, wait, let's pause and examine this more carefully.
        \item Wait a second, let's ensure this is right. Calculating carefully:
        \item Hmm, I want to verify this calculation. Let's go through it:
        \item Wait, this doesn't seem right. Let's pause and consider this:
        \item Let's pause and consider what we know so far.
        \item This didn't seem right. Wait, let's correct that.
        \item Wait, something seems off. Let's pause and consider what we know so far.
        \item Let's pause and consider if we've set up everything correctly.
        \item Wait a second. Is everything correct? Let me double-check.
        \item Wait, maybe there's something wrong. Let's pause and reconsider.
        \item The result looks strange, is everything correct? Let me double-check.
        \item Does this make sense? Let's rethink this.
        \item Could I have missed something? Let's pause and consider what we know so far.
        \item Wait $\star$
        \item The previous step is incorrect. $\star$
        \item There is a mistake.  $\star$
    \end{itemize}
    
    \textbf{\{Annotated Corrected Continuation\}} \\
    
    \end{tcolorbox}\caption{Dataset construction for SFSDPO with the set of transition phrases fetched from \citet{pan-etal-2025-lemma}. $\star$ denotes additional phrases.}
\label{fig:sfsdpo_construction} 
\end{figure*}

\begin{figure*}[h]
    \centering
    \includegraphics[width=0.8\linewidth]{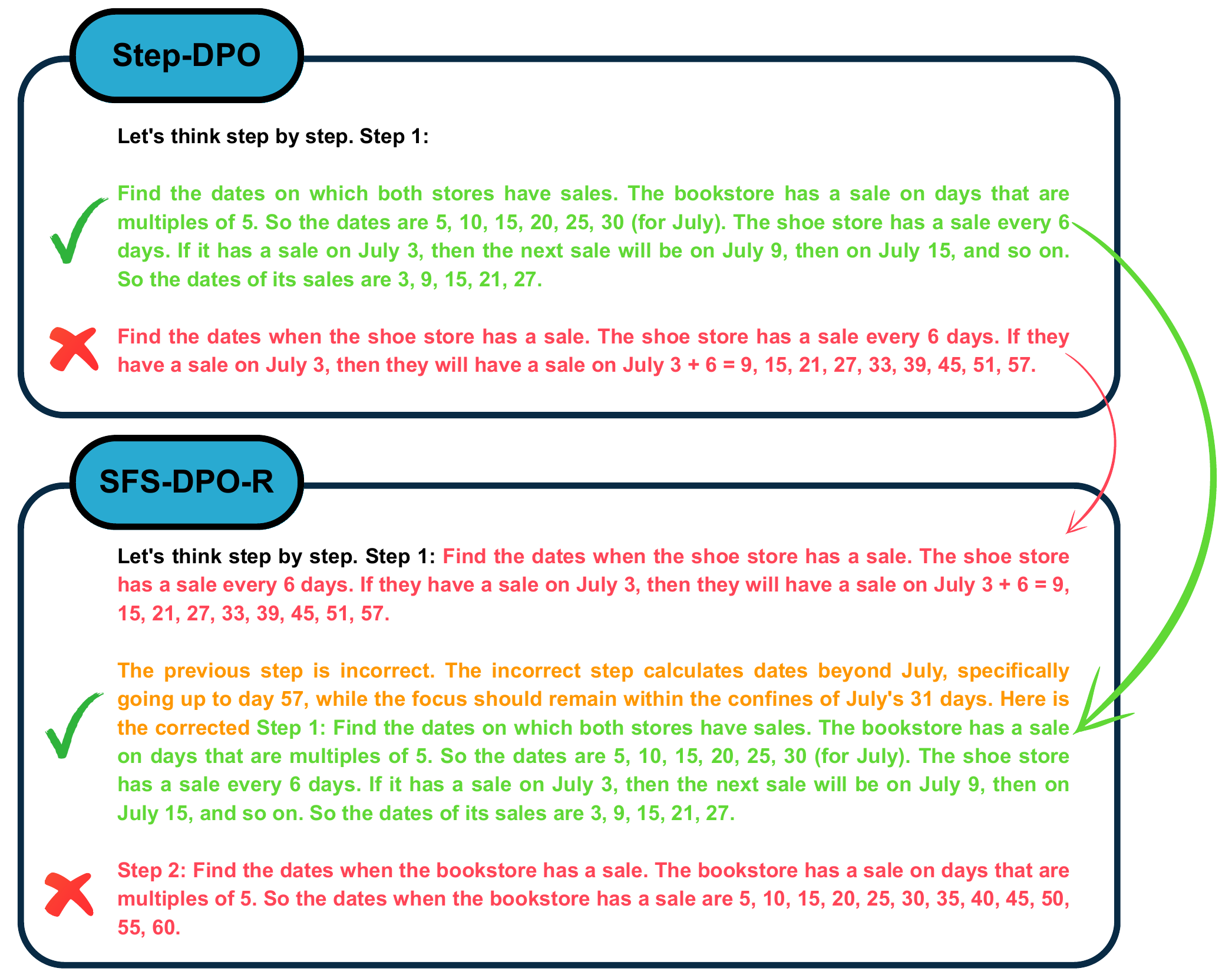}
    \caption{Illustration of the process for constructing the SFS-DPO-R dataset. Incorrect intermediate reasoning steps are identified and corrected with self-correction signals and self-correction reasonings.}
    \label{fig:sfsdpoexample}
\end{figure*}

\section{Qualitative Analysis of SFS-DPO-R Self-Correction}

The qualitative examples are provided in Tables~\ref{tab:sfsdpo_example_2}--\ref{tab:sfsdpo_example_8}. We include one comparison with LEMMA and S$^2$R, where these methods exhibit self-correction behavior but still produce excessive or insufficiently targeted revisions. The remaining examples focus on the base model and Step-DPO. The base model often propagates earlier errors without explicit repair, while Step-DPO can improve local step preference but still fails to revise already-generated incorrect steps. In contrast, SFS-DPO-R more reliably localizes erroneous intermediate steps and applies concise corrections while preserving coherent reasoning. These examples support that effective self-correction requires selective and accurate revision, not correction frequency alone.

\begin{table*}[t]
\centering
\renewcommand{\arraystretch}{1.15}
\setlength{\tabcolsep}{6pt}

\begin{tabular}{lp{13cm}}
\toprule
Problem & Courtney attended a concert and reported that the audience was 48 in number. However, Kelly went to the same concert and said that Courtney had made the mistake of overstating the number of people in attendance by 20\%. If Kelly was right, how many people really attended the concert?
\\
\midrule
Base Model
&  
Step 1: Identify the number of people Courtney reported.  
Courtney reported that the audience was 48 in number.

Step 2: Determine the percentage by which Courtney overestimated the number of people.  
According to Kelly, Courtney overestimated the number of people by 20\%.

\textcolor{red}{Step 3: Calculate the actual number of people based on the overestimation.  
If the number of people was overestimated by 20\%, then the actual number of people is 80\% of the number Courtney reported.  
To find 80\% of 48, we use the formula:
\[
\text{Actual number of people} = 48 \times \frac{80}{100} = 48 \times 0.8 = 38.4
\]
Since the number of people must be a whole number, we round 38.4 to the nearest whole number, which is 38. Therefore, the actual number of people who attended the concert is \boxed{38}.}
\\
\midrule 
\textbf{SFS-DPO-R} & Let's think step by step. Step 1: If Courtney's count of 48 was overstated by $20\%$, then the actual number of people at the concert is 48 minus $20\%$ of 48.\newline Step 2: To find $20\%$ of 48, we multiply 48 by $20/100$, which gives us $48 \times \frac{20}{100} = 9.6$.\newline \textcolor{red}{Step 3: To find the actual number of people at the concert, we subtract this 9.6 from the 48 that Courtney counted. So, the actual number of people is $48 - 9.6 = 38.4$. }S\newline\textcolor{orange}{However, the number of people cannot be a decimal, so there must be a mistake in our calculation. Let's retrace our steps and see where we went wrong. \newline}\textcolor{Green}{ Step 1: If Courtney's count of 48 was overstated by $20\%$, then the actual number of people at the concert is 48 minus $20\%$ of the actual number of people.\newline Let's call the actual number of people $x$. Then, $20\%$ of $x$ is $0.2x$. So, the equation becomes $48 = x + 0.2x$. \newline Step 2: Combining like terms on the right side gives $48 = 1.2 x$.\newline Step 3: To solve for $x$, we divide both sides of the equation by 1.2: $x = \frac{48}{1.2} = 40$. \newline So, the actual number of people at the concert is 40.
\newline The answer is: 40 }\\
\bottomrule
\end{tabular}
\caption{Qualitative reasoning between base Qwen2-7B-Instruct and Qwen2-7B-Instruct-SFS-DPO-R, where the base model produces a plausible but incorrect solution, while SFS-DPO-R identifies the error source and revises the reasoning to obtain the correct result.}
\label{tab:sfsdpo_example_2}
\end{table*}

\begin{table*}[t]
\centering
\small
\renewcommand{\arraystretch}{1.15} \setlength{\tabcolsep}{6pt} \begin{tabular}{lp{13cm}} \toprule Problem & Annie's soccer team is conducting a passing drill in which each of the 11 members of the soccer team must pass the ball to each of the other members exactly three times. How many times must the ball be passed before the passing drill is over? Please reason step by step, and put your final answer within \textbackslash boxed{}. \\ \midrule Step-DPO & \#\# Step 1: Determine the total number of passes each player must make Each of the 11 players must pass the ball to each of the other 10 players exactly three times. This means each player will make $10 * 3 = 30$ passes in total. \#\# Step 2: Calculate the total number of passes made by all players Since there are 11 players, and each player makes 30 passes, it might seem like the total number of passes is $11 * 30$. \textcolor{red}{However, this counts each pass twice (once for each player involved in the pass), so we need to divide this number by 2 to get the actual number of passes.} \textcolor{red}{\#\# Step 3: Calculate the total number of passes The total number of passes is $(11 * 30) / 2 = 165$.} \textcolor{red}{The final answer is: $\boxed{165}$} \\ \midrule 

S$^2$R & ... 

\textcolor{red}{The total number of unique passes is 330 / 2 = 165.}
\textcolor{orange}{Wait, let me recheck my solution. To verify the answer, we can consider the structure of the problem and the nature of the passing drill.} \textcolor{red}{ The problem involves 11 players, and each player must pass the ball to each of the other 10 players exactly three times. This means that each player will make 10 passes, and since there are 11 players, it might seem like there would be 11 x 10 = 110 passes. However, this counts each pass twice (once for each player involved in the pass), so we need to divide by 2 to get the correct number of unique passes. The correct calculation is (11 players x 10 passes/player) / 2 = 55 unique passes. Since each player must pass the ball 3 times to each of the other 10 players, we need to multiply the number of unique passes by 3 to get the total number of passes. 55 unique passes x 3 = 165 passes.The answer provided is 165, which matches the result of this reasoning. Therefore, the answer is correct} \\ \midrule

LEMMA & ... This means there are 55 unique pairs of players. Now, since each player must pass the ball to each of the other players three times, we need to multiply the number of unique pairs by the number of passes each player makes: \textcolor{red}{\[ \text{Total passes} = 3 \times 55 = 165.\]} \textcolor{orange}{Hmm, let's verify this calculation. Each of the 11 players is passing to 10 other players, and since each pass is counted three times, we can also express it as:} \textcolor{Green}{\[ \text{Total passes} = 3 \times (11 \times 10) = 3 \times 110 = 330. \]}\textcolor{orange}{Wait a second. This seems inconsistent with our earlier calculation. Let's clarify: we only need to count the unique pairs once, and then multiply by the number of passes per pair.} \textcolor{red}{So, the correct approach is indeed to calculate the total number of unique pairs and multiply by the number of passes per pair:} \textcolor{red}{ \[ \text{Total passes} = 3 \times 55 = 165. \]} \textcolor{red}{ Thus, the total number of times the ball must be passed before the drill is over is indeed \( \boxed{165} \).} \\ \midrule \textbf{SFS-DPO-R} & \#\# 

... 

\#\# Step 2: Calculate the total number of passes made by all players in one round of passing. If each of the 11 players passes to the other 10 players exactly 3 times, it might initially seem like the calculation is $11$ players $*$ $10$ players $* 3$ passes $= 330$ passes. \textcolor{red}{However, this counts each pass twice because when one player passes to another, it counts as a pass for both the passer and the receiver. So, we need to divide this number by 2 to get the actual number of unique passes in one round.} 

\#\# Step 3: Calculate the total number of passes needed for the drill. \textcolor{orange}{The calculation from Step 2, $11 * 10 * 3$, is the correct way to find the total number of passes needed because it correctly accounts for the fact that each of the 11 players must pass to the other 10 players exactly 3 times, without double-counting the passes.} \textcolor{Green}{So, the total number of passes needed is $11 * 10 * 3$ = 330.} \textcolor{Green}{ The final answer is: $\boxed{330}$}\\ \bottomrule \end{tabular}
\caption{Qualitative comparison on the MATH dataset between the base Llama-3.1-8B-Instruct-Step-DPO model and Llama-3.1-8B-Instruct-SFS-DPO-R. S$^2$R cannot perform correct self-correction while LEMMA exhibits excessive self-correction by repeatedly revising turning a correct reasoning path into an incorrect final answer, while SFS-DPO-R performs a targeted correction and reaches the correct answer.}

\label{tab:sfsdpo_example_3}
\end{table*}

\begin{table*}[t]
\centering
\small
\renewcommand{\arraystretch}{1.15}
\setlength{\tabcolsep}{6pt}

\begin{tabular}{lp{13cm}}
\toprule
Problem & Let $a,$ $b,$ $c,$ $d$ be nonnegative real numbers such that $a + b + c + d = 1.$  Find the maximum value of
\[a^2 + b^2 + c^2 + d^2.\]
\\
\midrule
Step-DPO
&  

... \#\# Step 2: We can use the Cauchy-Schwarz inequality to find a relationship between $a^2 + b^2 + c^2 + d^2$ and $a + b + c + d.$
Applying the Cauchy-Schwarz inequality to the numbers $a, b, c, d$ and $1, 1, 1, 1$, we get
\[(a^2 + b^2 + c^2 + d^2)(1^2 + 1^2 + 1^2 + 1^2) \ge (a + b + c + d)^2.\]
This simplifies to
\[a^2 + b^2 + c^2 + d^2 \le \frac{(a + b + c + d)^2}{4}.\]

\textcolor{red}{\#\# Step 3: Since we are given that $a + b + c + d = 1,$ we can substitute this into the inequality from the previous step to get
\[a^2 + b^2 + c^2 + d^2 \le \frac{1^2}{4} = \frac{1}{4}.\]
This means that the maximum value of $a^2 + b^2 + c^2 + d^2$ is $\frac{1}{4}.$}

\textcolor{red}{\#\# Step 4: To find when the maximum value is achieved, we need to find the values of $a, b, c, d$ that satisfy the equality condition of the Cauchy-Schwarz inequality.
The equality condition is satisfied when the two sequences are proportional, i.e., when $a = b = c = d = \frac{1}{4}.$}

\textcolor{red}{\#\# Step 5: Therefore, the maximum value of $a^2 + b^2 + c^2 + d^2$ is achieved when $a = b = c = d = \frac{1}{4}.$}

\textcolor{red}{The final answer is: $\boxed{\frac{1}{4}}$
}\newline

 \\
 \midrule
\textbf{SFS-DPO-R} & \#\# Step 1: Understand the problem and the constraint given
We are given that $a + b + c + d = 1$ and we need to find the maximum value of $a^2 + b^2 + c^2 + d^2.$ This is an optimization problem with a constraint.

\#\# Step 2: Identify a relevant inequality or method to apply
The Cauchy-Schwarz inequality is a powerful tool for problems involving sums of squares. It states that for any real numbers $x_1, x_2, \ldots, x_n$ and $y_1, y_2, \ldots, y_n,$
\[(x_1^2 + x_2^2 + \cdots + x_n^2)(y_1^2 + y_2^2 + \cdots + y_n^2) \ge (x_1y_1 + x_2y_2 + \cdots + x_ny_n)^2.\]
However, in this case, we can use a simpler inequality, the QM-AM (Quadratic Mean - Arithmetic Mean) inequality, which states that for nonnegative real numbers, the quadratic mean is greater than or equal to the arithmetic mean. That is,
\[\sqrt{\frac{a^2 + b^2 + c^2 + d^2}{4}} \ge \frac{a + b + c + d}{4}.\]
Squaring both sides of the QM-AM inequality and multiplying by 4 gives us
\[a^2 + b^2 + c^2 + d^2 \le \frac{(a + b + c + d)^2}{1} = (a + b + c + d)^2 = 1^2 = 1.\]
This inequality gives us an upper bound on $a^2 + b^2 + c^2 + d^2$.

\textcolor{red}{\#\# Step 3: Determine when equality occurs
To find the maximum value of $a^2 + b^2 + c^2 + d^2,$ we need to find when the equality occurs in the QM-AM inequality. Equality occurs when all the numbers are equal, i.e., $a = b = c = d = \frac{1}{4}.$}

\textcolor{red}{\#\# Step 4: Conclude the maximum value
Since the maximum value of $a^2 + b^2 + c^2 + d^2$ is achieved when $a = b = c = d = \frac{1}{4},$ the maximum value is
\[a^2 + b^2 + c^2 + d^2 = \left(\frac{1}{4}\right)^2 + \left(\frac{1}{4}\right)^2 + \left(\frac{1}{4}\right)^2 + \left(\frac{1}{4}\right)^2 = \frac{1}{16} + \frac{1}{16} + \frac{1}{16} + \frac{1}{16} = \frac{4}{16} = \frac{1}{4}.\]}
\textcolor{orange}{However, we can see that the previous step is incorrect. }\textcolor{Green}{The correct maximum value is simply the upper bound we found in Step 2, which is 1.}

\textcolor{Green}{The final answer is: $\boxed{1}$}\\
\bottomrule
\end{tabular}
\caption{Qualitative comparison on the MATH dataset between the base Llama-3.1-8B-Instruct-Step-DPO model and Llama-3.1-8B-Instruct-SFS-DPO-R. While Step-DPO arrives at an incorrect conclusion due to flawed inequality reasoning, SFS-DPO-R identifies the mistake, revises its reasoning, and recovers the correct maximum value.}

\label{tab:sfsdpo_example_4}
\end{table*}

\begin{table*}[t]
\centering
\renewcommand{\arraystretch}{1.15}
\setlength{\tabcolsep}{6pt}

\begin{tabular}{lp{13cm}}
\toprule
Problem & In a right triangle, the leg opposite angle x is 42 units long and the hypotenuse is 59 units long. Which of the following is NOT true: (A) $\sin x = \frac{42}{59}$ (B) $\sin(90^{\circ}-x)=\frac{42}{59}$ (C) $\cos(90^{\circ}-x)=\frac{42}{59}$ (D) $\sin(90^{\circ}-x)-cos=0$

\\
\midrule
Base Model
&  

Step 1: We know that in a right triangle, the sine of an angle is equal to the length of the opposite side divided by the length of the hypotenuse. So, $\sin x = \frac{42}{59}$. This means that statement A is true.

Step 2: We also know that $\sin(90^{\circ}-x)$ is equal to the cosine of the angle $x$. So, $\sin(90^{\circ}-x) = \cos x$. And we know that $\cos x = \frac{42}{59}$. So, $\sin(90^{\circ}-x) = \frac{42}{59}$. This means that statement B is true.

Step 3: We also know that $\cos(90^{\circ}-x)$ is equal to the sine of the angle $x$. So, $\cos(90^{\circ}-x) = \sin x$. And we know that $\sin x = \frac{42}{59}$. So, $\cos(90^{\circ}-x) = \frac{42}{59}$. This means that statement C is true.

Step 4: However, statement D says that $\sin(90^{\circ}-x) - \cos = 0$. But we just found out that $\sin(90^{\circ}-x) = \cos x$ and $\cos x = \frac{42}{59}$. So, $\sin(90^{\circ}-x) - \cos = \frac{42}{59} - \frac{42}{59} = 0$. This means that statement D is also true.

\textcolor{red}{<< Repeated block >> \newline
<< Repeated block >> \newline
<< Repeated block >> \newline
<< Repeated block >> \newline
No answer
}
 \\
 \midrule
\textbf{SFS-DPO-R} & Step 1: We know that in a right triangle, the sine of an angle is defined as the ratio of the length of the side opposite the angle to the length of the hypotenuse. So, $\sin x = \frac{42}{59}$, which means statement A is true.

Step 2: We also know that $\sin(90^{\circ}-x)$ is equal to the cosine of the angle $x$. This is a well-known identity in trigonometry. So, $\sin(90^{\circ}-x) = \cos x$. And we know that $\cos x = \frac{42}{59}$, which means statement B is true.

Step 3: We also know that $\cos(90^{\circ}-x)$ is equal to the sine of the angle $x$. This is another well-known identity in trigonometry. So, $\cos(90^{\circ}-x) = \sin x$. And we know that $\sin x = \frac{42}{59}$, which means statement C is true.

Step 4: However, statement D says $\sin(90^{\circ}-x)-cos=0$. But we just established that $\sin(90^{\circ}-x) = \cos x$ and $\cos x = \frac{42}{59}$. So, $\sin(90^{\circ}-x)-cos = \cos x - \cos x = 0$. This means statement D is true.

\textcolor{orange}{So, all the statements A, B, C, and D are true. But the question asks for the statement that is NOT true. So, there must be a mistake in our reasoning. Let's retrace our steps.}

\textcolor{Green}{Upon closer inspection, we realize that in step 4, we assumed that $\sin(90^{\circ}-x) = \cos x$. But this is not correct. The correct identity is $\sin(90^{\circ}-x) = \cos x$. So, statement B should be $\sin(90^{\circ}-x) = \cos x$, not $\sin(90^{\circ}-x) = \frac{42}{59}$.
}
\textcolor{Green}{With this correction, we can see that statement B is not true. So, the statement that is NOT true is B.
The answer is: B}\\
 \bottomrule
\end{tabular}
\caption{Qualitative comparison on the Gaokao2023 dataset between the base Qwen2-7B-SFT model and Qwen2-7B-SFT-SFS-DPO-R. While the base model produces a seemingly coherent but incomplete reasoning trajectory, SFS-DPO-R explicitly identifies the source of error at the step level and revises the flawed reasoning, ultimately leading to the correct final answer.}

\label{tab:sfsdpo_example_5}
\end{table*}

\begin{table*}[t]
\centering
\renewcommand{\arraystretch}{1.15}
\setlength{\tabcolsep}{6pt}

\begin{tabular}{lp{13cm}}
\toprule
Problem & A chord of a circle is perpendicular to a radius at the midpoint of the radius. The ratio of the area of the larger of the two regions into which the chord divides the circle to the smaller can be expressed in the form $\displaystyle {{a\pi+b\sqrt{c}}\over{d\pi-e\sqrt{f}}}$, where $a$, $b$, $c$, $d$, $e$, and $f$ are positive integers, $a$ and $e$ are relatively prime, and neither $c$ nor $f$ is divisible by the square of any prime. Find the remainder when the product $a\cdot b\cdot c\cdot d\cdot e\cdot f$ is divided by 1000.

\\
\midrule
Base Model
&  

Step 1: Draw a diagram.

Step 2:  Label the radius of the circle $r$.

Step 3:  Draw the chord and the two radii that are perpendicular to the chord.

...

Step 11:  Find the product $a\cdot
b\cdot c\cdot d\cdot e\cdot f$.

Step 12:  Find the remainder when the product is divided by 1000. The final answer is $\boxed{120}$.
 \\
 \midrule
\textbf{SFS-DPO-R} & Step 1: Given a circle with radius \( R \), a chord is perpendicular to a radius at the midpoint of the radius. This means the distance from the center of the circle to the chord is \( \frac{R}{2} \). The length of the chord can be found using the Pythagorean theorem in the right triangle formed by the radius, half the chord, and the distance from the center to the chord. The length of half the chord is \( \sqrt{R^2 - \left(\frac{R}{2}\right)^2} = \sqrt{R^2 - \frac{R^2}{4}} = \sqrt{\frac{3R^2}{4}} = \frac{R\sqrt{3}}{2} \). Therefore, the length of the chord is \( R\sqrt{3} \).

\textcolor{red}{Step 2: The area of the circle is \( \pi R^2 \). To find the area of the smaller region formed by the chord, we need to calculate the area of the circular segment. The angle subtended by the chord at the center of the circle can be found using the cosine rule in the triangle formed by two radii and the chord. The cosine of the angle \( \theta \) is given by \( \cos \theta = \frac{\left(\frac{R}{2}\right)^2 + \left(\frac{R}{2}\right)^2 - R^2}{2 \cdot \frac{R}{2} \cdot \frac{R}{2}} = \frac{\frac{R^2}{4} + \frac{R^2}{4} - R^2}{\frac{R^2}{2}} = \frac{\frac{R^2}{2} - R^2}{\frac{R^2}{2}} = \frac{-\frac{R^2}{2}}{\frac{R^2}{2}} = -1 \).} \textcolor{orange}{However, this is incorrect because we need to consider the angle subtended by the chord on the circle, which is \( 2\theta = 2 \cos^{-1}\left(-\frac{1}{2}\right) = 2 \cdot \frac{2\pi}{3} = \frac{2\pi}{3} \).}

\textcolor{Green}{Step 2: The area of the sector with angle \( \frac{2\pi}{3} \) is \( \frac{1}{3} \pi R^2 \). The area of the triangle formed by the two radii and the chord is \( \frac{1}{2} R^2 \sin \left(\frac{2\pi}{3}\right) = \frac{1}{2} R^2 \cdot \frac{\sqrt{3}}{2} = \frac{R^2 \sqrt{3}}{4} \). Therefore, the area of the smaller segment is \( \frac{1}{3} \pi R^2 - \frac{R^2 \sqrt{3}}{4} = R^2 \left( \frac{\pi}{3} - \frac{\sqrt{3}}{4} \right) \).}

...

Step 6: To find the remainder when the product \( a \cdot b \cdot c \cdot d \cdot e \cdot f \) is divided by 1000, we identify \( a = 8 \), \( b = 3 \), \( c = 3 \), \( d = 4 \), \( e = 3 \), and \( f = 3 \). The product is \( 8 \cdot 3 \cdot 3 \cdot 4 \cdot 3 \cdot 3 = 2592 \). The remainder when 2592 is divided by 1000 is \( 592 \).

Thus, the answer is \( \boxed{592} \).

\\
 \bottomrule
\end{tabular}
\caption{Qualitative comparison on the MATH dataset between the base Qwen2.5-Math-7B-Instruct model and Qwen2.5-Math-7B-Instruct-SFS-DPO-R. SFS-DPO-R demonstrates explicit self-correction by detecting and revising an incorrect intermediate reasoning step, ultimately recovering the correct solution, while the base model produces an incorrect answer without correction.}

\label{tab:sfsdpo_example_6}
\end{table*}

\begin{table*}[t]
\centering
\renewcommand{\arraystretch}{1.15}
\setlength{\tabcolsep}{6pt}

\begin{tabular}{lp{13cm}}
\toprule
Problem & A regular octagon $ABCDEFGH$ has an area of one square unit. What is the area of the rectangle $ABEF$? [asy] pair A,B,C,D,I,F,G,H; A=(-1,1); B=(1,1); H=(-2.41,-0.41); C=(2.41,-0.41); G=(-2.41,-2.41); D=(2.41,-2.41); F=(-1,-3.82); I=(1,-3.82); draw(A--B--C--D--I--F--G--H--cycle,linewidth(0.7)); label("$A$",A,N); label("$B$",B,N); label("$C$",C,E); label("$D$",D,E); label("$E$",I,S); label("$F$",F,S); label("$G$",G,W); label("$H$",H,W); [/asy]

\\
\midrule
Base Model
&  

Step 1: Understand the structure of the regular octagon and the rectangle $ABEF$.

A regular octagon can be divided into 8 isosceles triangles, each with a central angle of \(45^\circ\). The area of the octagon is given as 1 square unit.

...

\textcolor{red}{Step 4: Calculate the area of the rectangle $ABEF$.}

\textcolor{red}{Since each triangle has an area of \(\frac{1}{8}\) square units, the area of the rectangle $ABEF$ (which is composed of two such triangles) is:
\[
\text{Area of } ABEF = 2 \times \frac{1}{8} = \frac{2}{8} = \frac{1}{4} \text{ square units}
\]}

\textcolor{red}{Thus, the area of the rectangle $ABEF$ is:
\[
\boxed{\frac{1}{4}}
\]}
 \\
 \midrule
\textbf{SFS-DPO-R} & 

Step 1: Understanding the Geometry of the Regular Octagon
A regular octagon can be divided into 8 isosceles triangles by drawing lines from the center to each vertex. Each of these triangles has a central angle of \( \frac{360^\circ}{8} = 45^\circ \).

...

\textcolor{red}{Step 5: Calculating the Area of the Rectangle \(ABEF\)
The rectangle \(ABEF\) can be seen as composed of two isosceles triangles and a central square. Each isosceles triangle has an area of \( \frac{1}{8} \) of the octagon's area, which is \( \frac{1}{8} \) square units. Since there are two such triangles in the rectangle, their combined area is \( 2 \times \frac{1}{8} = \frac{1}{4} \) square units.}

 \textcolor{Orange}{The central square, which is part of the rectangle, has an area equal to the remaining part of the octagon after removing the 8 isosceles triangles. The area of the central square is \( 1 - 8 \times \frac{1}{8} = 1 - 1 = 0 \) (this is incorrect, so we need to re-evaluate).}

\textcolor{Green}{The correct approach is to recognize that the central square's area is the remaining part after removing the 4 pairs of triangles, which is \( 1 - 4 \times \frac{1}{4} = 1 - \frac{1}{2} = \frac{1}{2} \).}

\textcolor{Green}{Thus, the area of the rectangle \(ABEF\) is the sum of the areas of the two isosceles triangles and the central square:
\[ \text{Area of } ABEF = \frac{1}{4} + \frac{1}{2} = \frac{1}{4} + \frac{2}{4} = \frac{3}{4} \]}

\textcolor{Green}{Therefore, the area of the rectangle \(ABEF\) is \(\boxed{\frac{1}{2}}\).}
\newline
\\

 \bottomrule
\end{tabular}
\caption{Qualitative comparison on the MATH dataset between the base Qwen2.5-14B-Instruct model and Qwen2.5-14B-Instruct-SFS-DPO-R. SFS-DPO-R exhibits self-correction by identifying that the initial decomposition of rectangle ABEF is incomplete, but its revised reasoning remains internally inconsistent, leading to an incorrect final answer despite attempting correction.}

\label{tab:sfsdpo_example_7}
\end{table*}

\begin{table*}[t]
\centering
\renewcommand{\arraystretch}{1.15}
\setlength{\tabcolsep}{6pt}
\begin{tabular}{lp{13cm}}
\toprule
Problem &
Patrick has three glue sticks that are partially used. One has
$1/6$ left, the second has $2/3$ left, and the third one has $1/2$ left.
If a glue stick is 12 millimeters long originally, what is the total length
of the glue sticks that are not used?  \\
\midrule

Step-DPO &
Step 1: Interpret the given fractions.
\textcolor{red}{The response incorrectly treats the fractions left as if
they should be converted into fractions already used. It therefore claims
that the first, second, and third glue sticks have $5/6$, $1/3$, and $1/2$
used, respectively.}

Step 2: Calculate the lengths.
\textcolor{red}{It calculates
$12\left(\frac{5}{6}\right)=10$ mm,
$12\left(\frac{1}{3}\right)=4$ mm, and
$12\left(\frac{1}{2}\right)=6$ mm.}

Step 3: Return the result.
\textcolor{red}{The total used length is
$10+4+6=20$ mm. The final answer is $\boxed{20}$.} \\
\midrule

\textbf{SFS-DPO-R} &
Step 1: Interpret ``not used.''
The response initially considers computing the unused length using the
complement of a used fraction.

\textcolor{orange}{But wait---the question says that the glue sticks have
$1/6$, $2/3$, and $1/2$ left. Therefore, the fractions left are already the
unused portions; they should not be converted into their complements.}

Step 2: Correct the calculation.
\textcolor{Green}{The unused lengths are
$12\left(\frac{1}{6}\right)=2$ mm,
$12\left(\frac{2}{3}\right)=8$ mm, and
$12\left(\frac{1}{2}\right)=6$ mm.}

Step 3: Return the corrected result.
\textcolor{Green}{Thus, the total unused length is
$2+8+6=16$ mm, and the final answer is $\boxed{16}$.} \\
\bottomrule
\end{tabular}

\caption{Qualitative comparison on GSM8K between Qwen3-8B-Step-DPO and
Qwen3-8B-SFS-DPO-R. Step-DPO mistakes the fractions left for fractions used and returns
incorrect answer, whereas SFS-DPO-R identifies the interpretation error and correctly
returns groundtruth answer.}
\label{tab:sfsdpo_example_8}
\end{table*}
\label{sec:appendix}

\end{document}